# Intelligent Energy Management in Smart Cities: Leveraging IoT and Machine Learning for Optimizing Complex Networks and Systems

Maryam Nikpour [a], Parisa Behvand Yousefi [b], Hadi Jafarzadeh [c], Kasra Danesh[d], Roya Shomali [e], Saeed Asadi[f], Ahmad Gholizadeh Lonbar[g], Mohsen Ahmadi[d,*]

*[a] Architecture Department, Ahvaz Branch, Islamic Azad University, Ahvaz, Iran*
*[b] School of E-Learning, Khaje Nasirodin Toos (K.N.Toosi) university of technology, Tehran, Iran*
*[c] School of E-Learning, Shiraz University, Shiraz, Iran*
*[d] Department of Electrical Engineering and Computer Science, Florida Atlantic University, Boca Raton, Florida, USA*
*[e] Department of Information Systems, Statistics and Management Science, The University of Alabama, Tuscaloosa, USA*
*[f] Department of Civil Engineering, The University of Texas at Arlington, Arlington, Texas, USA*
*[g] Department of Civil, Construction, and Environmental Engineering, University of Alabama, Tuscaloosa, AL, USA*
***Corresponding author:** *Mohsen.Ahmadi (Email: mahmadi2021@fau.edu)*

## Abstract

This study addresses the growing challenges of energy consumption and the depletion of energy resources, particularly in the context of smart buildings. As the demand for energy increases alongside the need for efficient building maintenance, it becomes imperative to explore innovative energy management solutions. We present a review of Internet of Things (IoT)-based frameworks aimed at managing smart city energy consumption, the pivotal role of IoT devices in addressing these issues due to their compactness, sensing, measurement, and computing capabilities. Our review methodology involves a thorough analysis of existing literature on IoT architectures and frameworks for intelligent energy management applications. We focus on systems that not only collect and store data but also support intelligent analysis for monitoring, controlling, and enhancing system efficiency. Additionally, we examine the potential for these frameworks to serve as platforms for the development of third-party applications, thereby extending their utility and adaptability. The findings from our review indicate that IoT-based frameworks offer potential to reduce energy consumption and environmental impact in smart buildings. By adopting intelligent mechanisms and solutions, these frameworks facilitate effective energy management, leading to improved system efficiency and sustainability. Considering these findings, we recommend further exploration and adoption of IoT-based wireless sensing systems in smart buildings as a strategic approach to energy management. Our review highlights the importance of incorporating intelligent analysis and enabling the development of third-party applications within the IoT framework to efficiently meet evolving energy demands and maintenance challenges.

## 1- Introduction

The rapid increase in global energy demand, driven by population growth and urbanization, presents an urgent challenge for sustainable development. As cities expand and energy consumption soars, efficient energy management becomes critical to mitigating environmental impacts and ensuring reliable energy supply. In this context, smart cities offer a promising solution, leveraging advanced technologies such as the Internet of Things (IoT) to automate energy production, distribution, and consumption. However, the existing energy infrastructure is often inefficient, leading to significant waste and resource depletion. Addressing these inefficiencies through intelligent, IoT-based frameworks is vital for achieving sustainability and energy resilience in the modern urban landscape. Despite the growing need for energy for a nation's economic development, global population growth, as well as for society's way of life, the worldwide energy demand has risen substantially [1]. The use of wide-area monitoring and smart meters has enabled automation in electrical energy production, transmission, distribution, and consumption at multiple levels. By measuring the electrical properties of the transmission system and utilizing universal timestamps, it becomes possible to accurately predict measurement precision, identify faults, and isolate them effectively. This enhances the reliability of the transmission grid. However, grid distribution and usage differ in terms of the level of automation. Advancements in the IoT have played a crucial role in further automating these processes and controlling systems more efficiently [3]. Sustainability is the use of energy efficiently while meeting the demands of the modern world. Ecological, social, and economic concerns can all be addressed through sustainable energy. Sustainable energy management may involve utilizing renewable resources in the best possible way, leveraging sustainable energy resources, utilizing alternative energy resources, integrating IoT solutions, and applying green technology. Incorporating new technologies into existing operations is the goal of a smart system. Smart systems extract information about a city's utility system and traffic flow using sensors. These systems analyze information patterns for forecasting. An IoT system sends data through a network without requiring human-to-human interaction. Green technology refers to technologies that build environmentally friendly products by combining science and technology. In addition to purifying water, preserving natural resources, producing clean energy, and recycling garbage, green technologies are also used. Interconnected digital and mechanical machinery, things, and computer devices also use this technology.

Sustainable energy systems include intelligent grids, intelligent cities, and intelligent transportation systems. Implementing a sustainable energy system is not limited to large-scale techniques; emailing and texting without using paper can also reduce paper consumption. Both of these contribute to sustainable energy development. It is a formidable task for the government to generate and distribute energy to a growing population. Energy consumption has increased due to a growing population, resulting in several environmental and health concerns. Therefore, creating and utilizing energy resources effectively is crucial. The implementation of a sustainable energy system entails three primary phases, according to Goncalves and Santos [1]. They are 1) to analyze energy management policies at the industrial level; 2) to debate ways to implement sustainable energy; and 3) to examine the best ways to use sustainable energy once it has been validated by international experts [2]. Recent years have seen dramatic growth in the number of smart gadgets, which is soon likely to surpass billions. This is owing to the proliferation of IoT technologies across all industries. In smart networks, there are mainly IoT systems composed of networked devices, such as smartphones, sensors, automobiles, and so on. As part of sustainable energy management, smart grids are one of the solutions. This can be explained in part by introducing smart, autonomous, and bidirectional power grids [3]. Threatening the environment and working against green technology and carbon emission reductions, these tactics have a negative effect on the environment. They are designed mainly for the benefit of corporations and governments. Historically, most energy meters have been analog. Errors, operational losses, and theft were common. Most of the time, the defect cannot be diagnosed properly, and it cannot be corrected. As a result, the smart meter facilitated communication between the meter, the grid, and the user [4]. The expansion of macro data and the evolution of IoT technology play a crucial role in making smart cities possible. Using high-end network services, metadata can provide valuable insights into large volumes of data collected from a variety of sources, and the IoT has enabled sensor integration, radio frequency, and Bluetooth detection in the real world. Combining metadata with the IoT is an unexplored research area that can be used to reach the smart city of the future while presenting new and exciting challenges [5-7]. In smart cities, several technologies are used to increase citizen comfort levels in areas such as transportation, energy, education, etc. This also includes reducing costs and using resources in addition to promoting active and effective citizen participation [8]. A recent technology that plays an important role in increasing smart city services is metadata analysis. In smart cities, smart buildings, smart healthcare, and smart industrial applications, IoT use BD technology to deploy tiny sensors for wireless communication. The need for a definition of building energy use grows as more projects and technologies address this issue [9]. It is necessary to evaluate how a building functions to determine how much energy it consumes. The energy consumption of residential buildings is mainly affected by the indoor comfort services provided to individuals, whereas the energy consumption of industrial buildings is largely determined by industrial machinery and manufacturing infrastructure. Integrating and developing systems based on Information and Communication Technologies, and, the IoT, facilitates a wide array of applications and makes smart buildings a reality [10, 11]. Data and knowledge from the real world are effectively absorbed into the digital realm through IoT and communication between intelligent objects. Their study evaluates the effectiveness of IoT-based frameworks in managing energy within smart cities, specifically evaluating the impact of these technologies on reducing energy consumption and reducing environmental impacts in smart buildings. They aim to identify and advocate for solutions that enhance system efficiency and sustainability by reviewing IoT architectures and frameworks designed for intelligent energy management. This study highlights key challenges, such as the interoperability of IoT devices, data security, and the scalability of systems across diverse urban environments. Our original contributions include the development of a framework for integrating third-party applications, enabling adaptive energy management in smart cities. We address the gaps in existing literature by proposing solutions that not only optimize energy usage but also ensure system flexibility and resilience in evolving energy landscapes. It focuses on the implementation of advanced algorithms and technologies that are capable of autonomously analyzing, predicting, and managing energy consumption in smart buildings and cities. Using machine learning, artificial intelligence (AI), and the IoT, these mechanisms are capable of monitoring and controlling energy consumption to optimize efficiency and reduce waste. Incorporating ML into energy management enables dynamic adjustments to supply and demand, reducing energy waste while ensuring a steady and reliable power supply. As these algorithms learn from historical data, they can detect anomalies and automatically adjust to changing conditions, such as variations in user behavior or environmental factors. Additionally, ML facilitates the integration of renewable energy sources by predicting their availability and incorporating them into the grid efficiently. In smart buildings, ML systems can autonomously control heating, cooling, and lighting based on real-time data from sensors, further enhancing energy efficiency. These advanced systems minimize human intervention by autonomously monitoring and adjusting energy usage, leading to cost reductions and a smaller environmental footprint. Ultimately, ML-powered IoT frameworks foster smarter, more sustainable urban energy systems, contributing to overall energy efficiency and environmental conservation. These solutions are capable of forecasting energy demand, adapting to user behavior, and seamlessly integrating renewable energy sources through the integration of data from various sensors and devices. As a result of this approach, not only do energy systems operate more efficiently, but also the environmental impact of energy consumption is minimized. In the field of energy management, the solutions are unique in that they are capable of learning from data, making informed decisions, and dynamically adapting to changing conditions without human intervention.

The availability of smart gadgets with sensing and interaction capabilities, as well as recognition technologies, enables the collection of unprecedented amounts of actual environmental information. As described in this article, we propose an infrastructure for a smart city that is based on the concept of the digital city and integrates the IoT to control energy consumption. Three different methods were used to analyze the data in this study. To examine the various components of the smart grid (where such solutions (and results) can be applied, with a focus on smart city energy management), applications in the field of IoT and nearby are examined. After reviewing the research literature, a generalized set of existing solutions is developed; these include the key features identified during the literature review. Our study evaluates the effectiveness of IoT-based frameworks in managing energy within smart cities, specifically evaluating the impact of these technologies on reducing energy consumption and reducing environmental impacts in smart buildings. We aim to identify and advocate for solutions that enhance system efficiency and sustainability as well as support the integration and development of third-party applications to address evolving energy demands by reviewing IoT architectures and frameworks designed for intelligent energy management.

This paper makes several key contributions to the field of smart city energy management through the application of IoT frameworks:

- We provide a detailed examination of existing IoT-based frameworks for energy management, advancements and identifying gaps in current technologies.
- Our analysis demonstrates how IoT devices can enhance energy efficiency by leveraging intelligent analysis and real-time data processing.
- We explore the potential for IoT frameworks to support third-party applications, broadening their utility and adaptability in urban settings.
- Based on our findings, we offer targeted recommendations for future research and practical implementation of IoT solutions in smart buildings.

The Introduction section outlines the significance of sustainable energy management in smart cities, emphasizing the role of IoT-based frameworks in optimizing energy consumption, minimizing environmental impact, and enhancing system efficiency. The Literature Review provides a systematic analysis of recent studies on IoT applications in energy management, advancements in IoT architectures, energy optimization techniques, and machine learning integration while identifying critical gaps. The Components of Smart Cities section explores the foundational elements of smart cities, including data collection, processing, and application layers, along with key technologies enabling their development. The Discussion section critically evaluates the potential and challenges of IoT-based frameworks, addressing technical, economic, and social considerations. The Conclusion summarizes the key findings, presents actionable recommendations for future research, and highlights the limitations of the study. Finally, the paper offers Future Directions to guide researchers and stakeholders in advancing IoT-enabled energy management systems.

## 2- Literature Review

In our literature review, we applied a systematic approach to ensure a rigorous analysis of relevant studies. We began by identifying key search terms related to IoT, energy management, smart cities, and sustainability. Databases such as IEEE Xplore, ScienceDirect, and Google Scholar were searched using these terms. Our inclusion criteria focused on peer-reviewed articles published in the last decade, prioritizing studies that specifically addressed IoT applications in energy management within urban environments. We also considered papers that proposed practical frameworks or provided empirical data on energy efficiency and sustainability in smart buildings.

Studies were selected based on their relevance, methodological soundness, and contribution to the field. To ensure objectivity, the selected literature was analyzed using a structured framework that categorized studies by their focus areas, including IoT architectures, energy optimization techniques, machine learning applications, and sustainability impacts. Justifying these choices, our goal was to capture a broad spectrum of current advancements while also identifying critical gaps in the existing body of research. This process strengthened the rigor of our review by ensuring that the most relevant and high-impact studies were included for analysis.

There is a detailed discussion in this section of prior studies on IoT-based energy management in smart buildings, exploring various frameworks and their contributions to reducing energy consumption, with references to key works reported in the field. Zakik et al [12] in a study entitled Machine Learning System based on public sector energy efficiency management as a smart city approach. The purpose of this paper is to address the question of how Big Data operating systems and machine learning can be integrated into an intelligent system for managing public sector energy efficiency as an integral part of smart city concepts. Wang et al. [13] conducted a case study and developed optimization algorithms for a home energy management system based on the IoT. To demonstrate the value of the proposed technique, the results of multi-objective optimization are compared with those of Particle swarm optimization (PSO)-based and Bayesian Optimization Algorithm (BOA)-based algorithms. To evaluate system performance, the simulation results are compared with those of a normal home energy management system. In their study, Sadeeq et al. [14] considered energy management within IoT systems. A review of the literature and policy analysis related to aggregators of energy management systems and end users in affiliate systems is discussed in this article. Silva et al. [15] conducted a study entitled Using Big Data Analysis and Smart Web Architecture to Provide Efficient Services and Energy Management. To improve the network performance of a proposed smart city design, researchers have created an IoT and Web of Things integrated smart building architecture (smart house). In a study entitled Object-Based Energy Management, Shafik et al. [16] addressed the challenges and solutions associated with smart cities. An in-depth analysis of the proper management of the object-based Internet in smart cities is presented in this study. In this study, it is demonstrated that the IoT has increased energy consumption, and the summary study illustrates the latest proposed methods in energy management across different belts such as smart homes, smart buildings, and smart grids.

In addition, Jiang et al. [17] conducted a study on IoT-based smart city information system building and cloud computing. This study illustrates that this system can share information sharing, exchange, and fusion between different sensing subsystems, thereby solving the problem of previous information islands and serving the real needs of smart cities. According to Golpira et al. [18], a study called Optimized Urban Intelligent Management based on IoT while accounting for movable loads and energy savings was conducted. An IoT-based energy management framework for IoT-based public distribution networks is presented in this paper using a complex integer linear programming problem. Golpira et al. [19] conducted a study titled Optimized Urban Intelligent Management based on the IoT for Movable Loads and Energy Savings. An IoT-based energy management framework is proposed for IoT-based public distribution networks in smart cities (SC) by posing a new complex integer linear programming problem. Naranjo et al. [20] investigated City Smart Network Architecture using Fog for managing IoT applications. According to the simulation results of the selected case study, the energy-efficient solution provided by FOCAN impacts the communication performance of different objects within smart cities. Sodhro et al. [21] proposed solutions to address resource constraints in IoT-enabled devices, critical for sustainable smart cities. They introduced the Hybrid Adaptive Bandwidth and Power Algorithm (HABPA) and Delay-tolerant Streaming Algorithm (DSA) to enhance multimedia transmission performance. Using the video stream StarWarsIV, their analysis focused on metrics like power usage, battery life, delay, standard deviation, and packet loss ratio (PLR). HABPA showed better results than DSA and a baseline model, reducing power drain by 45%, improving battery life by 37%, and achieving a lower PLR of 4.5% and standard deviation of 3.5 dB. These findings demonstrate HABPA's potential to optimize energy efficiency and resource use in IoT devices, supporting smarter, more sustainable city operations. Kabalchi et al. [22] examined IoT applications for intelligent grids and smart environments. Additionally, the authors discussed obstacles, unanswered questions, and future research prospects for IoT-based EI applications, as well as several crucial study areas. Creating an Intelligent and Sustainable Environment Using the IoT was the title of a study conducted by Tyagi et al. [23]. Due to rapid technological advancements (in the last two decades), the Internet has improved the means/means of human life.

Today, smart cities or environments are essential for any government, anywhere in the world. An IoT network based on blockchain-based secure device management has been published by Gang et al. [24]. As a result of this article's framework, devices will be able to be managed confidentially and reliably within a smart city network, and the integrity, controllability, compatibility, and validation of each device will be provided, along with an analysis of the framework's efficacy. Their research draws upon these foundational works to propose a unique integration strategy that leverages IoT for real-time, adaptive energy management, setting the stage for a scalable smart city framework. This strategic approach aims to resolve the practical and operational inefficiencies identified in previous studies, thereby catalyzing the evolution from theoretical models to actionable, impactful urban energy solutions.

Yu et al. [25] developed a smart energy management system based on IoT in a city. Results show that the IoT-based intelligent city energy management system can satisfy the "smart city" construction needs. It integrates isolated energy information and measures energy consumption. Based on this, energy savings in urban energy consumption can be identified and evaluated. A study conducted by Javid et al. [26] is titled The General Architecture of the Internet to Control Electricity Consumption in Smart Homes. By using the Hadoop ecosystem, data is also processed in real-time to maximize productivity and minimize processing time. Alavi et al. [27] published a study titled Smart Cities equipped with the IoT: The most advanced and future trends. In addition, an IoT-based prototype is presented to illustrate how civilian infrastructure can be monitored in real-time using a cost-effective IoT-based prototype. Their final section discusses the challenges and future directions of IoT-based smart city applications. According to Zhang et al. [28] big data analytics can be used to develop smart city infrastructure. This study proposes measures, including rational planning of city infrastructure, the creation and improvement of performance mechanisms, and city management performance. Based on the results, the proposed system is more scalable and efficient than existing systems. In addition, the system performance is measured in terms of performance and processing time. Barto et al. [29] presented SIGINURB, an IoT-based smart city system. A hybrid system using cloud data processing analysis for the development of smart cities and urban planning is proposed in this study to address the challenges associated with smart cities. The purpose of the system is to create a variety of applications that provide new services to students, employees, companies, and government managers at the University of So Paulo, and improve their quality of life. Liu et al. [30] Designed an IoT-based energy management system based on cloud computing infrastructure in a smart city. An overview of Internet-based energy management in smart cities is provided in this study. then propose a framework and software model for an Internet-based system with cloud computing. Finally, the researchers suggest a planning scheme that is energy efficient. They also demonstrate its effectiveness.

According to Ejaz et al. [31], the increase in car use calls for innovative solutions to improve living conditions and transportation. The IoT connects diverse devices and systems to create intelligent cities. As the number and needs of IoT devices continue to grow, so does their energy consumption. Therefore, intelligent city solutions must be able to use energy efficiently and address the associated challenges. Extending the lifespan of low-power products through intelligent city energy harvesting is becoming increasingly popular. Initially, this approach focuses on programming energy-efficient smart homes, followed by wireless power distribution in smart cities for the IoT. However, a drawback is the high cost of power. In the past decade, real-time data sets were applied to traditional regression models [32], [33], with forecasts recorded afterward. Neural networks were subsequently used for prediction, as described in [34]. In this project, a simulation route mimics multi-layered feedforward networks, and the system's performance has been reported to be exceptional. In an extensive study on IoT, Reka and Dragicevic [35] found that intelligent grid techniques could improve energy efficiency. IoT offers numerous advantages for energy management, based on factors such as energy efficiency, security, functionality, and sustainability [36]. Asgari et al. [118] analyze sustainable welfare dynamics using GDP threshold effects and Granger causality. The study reveals that short-term economic growth drives energy consumption under certain GDP thresholds, while long-term impacts vary, the need for differentiated economic strategies based on GDP levels to achieve sustainability goals.

In [37] developed a method for analyzing time series load data by using auto-regressive integrated moving averages. Using forecasting, this effort seeks to reduce the peak-to-average ratio so that an intelligent IoT setup can be managed directly. To anticipate non-seasonal and seasonal data loads, [38] suggests using an Auto-Regressive Integrated Moving Average. In [39] developed a technique for forecasting short-term load. This is a data-driven design derived from a belief network with copula stages. The deep neural network technique and extreme learning machines are interrelated. In [40] developed an improved short-term prediction model based on deep residual networks. To improve the effectiveness of the system, a two-stage ensemble approach was developed. In [41] presented an Accurate and Fast Converging Short-Term Load Forecasting setup that includes a predictor, an optimizer, and a feature selector. It is available for common industrial SG applications, including load switching, infrastructure maintenance, contract evaluation, energy production planning, and power procurement. Short-term load forecasts are required in the electric sector of the energy market. Electric utilities use these models as part of their daily operations for evaluating contracts, performing infrastructure maintenance, planning power production, and shifting loads. Time series data may be selected based on various features [42,43]. All data sets are purged of duplicates before being input into the neural network that anticipates the contraction of electricity in the next few days. The authors also provide a way to overcome numerous hurdles and difficulties in load prediction setup [44]. To categorize this issue, we discover multiple hybrid techniques based on Computational intelligence. As opposed to Computational intelligence load prediction techniques, which are structured methodological approaches intended for prediction situations, the associated technology is a specific model that, along with other associated models such as regression and neural networks, is grouped into one technological category. Li et al. [45] developed an IoT-based technique for load prediction. Next, a two-step technique of prediction was developed, which provides more accurate predictions than conventional methods. In [46, 47] designed a hybrid model for analyzing massive amounts of data for use in a variety of studies. In [48] developed a method to improve voltage stability using machine learning. For optimal results, both Naive Bayes and K-Nearest Neighbor classifiers should be used. Rafiei et al. [49] developed a hybrid approach that integrates probabilistic load prediction with massive learning machines. The proposed system divides electricity expansion into a series of well-behaved and productive subseries using wavelet transformation. An optimization strategy based on swarms was proposed in [50] to predict short-term load. The term Darwinian particle swarm optimization extreme machine learning refers to Darwinian Particle Swarm Optimization. With this method, each swarm functions independently, just like in a normal PSO, to simulate natural selection. ELMs are among the most prevalent hidden layer feedforward networks. Compared to standard artificial learning approaches, they offer greater generalization. By eliminating the unnecessary hidden nodes and training difficulties, the selected route can give excellent results. In [8], the energy renewal of current building practices was evaluated as a cornerstone of the Italian energy strategy. The primary objective of the system is to conserve energy and ensure the effective use of resources through optimized ventilation, air conditioning, and heating. The research focuses on analyzing energy consumption, Indoor Environmental Quality, and the impact of varying airflow rates, utilizing both in situ measurements and dynamic simulations. While the system reduces energy usage, it is noted that it may increase carbon dioxide levels by up to 42% [51]. Yağ et al. [117] proposed intelligent energy management solutions that employ sophisticated mechanisms, such as machine learning and artificial intelligence, to maximize energy efficiency and sustainability in smart environments. In the study, efficiency encompasses multiple dimensions of IoT-based energy management in smart cities: optimizing energy utilization to minimize waste, improving operational performance to automate and enhance system responsiveness, reducing energy costs through effective demand and peak load management, and minimizing environmental impacts by lowering emissions associated with energy consumption. Efficiency also covers the optimal allocation of resources, ensuring that both technological and human assets are utilized effectively to achieve sustainable urban development. The related work based on the last study is shown in Table 1.

**Table 1.** Related work comparison.

| Authors | Goal | Protocol | Efficiency | Surveillance environment | Security | Speed of transmission | Service quality |
|---|---|---|---|---|---|---|---|
| [52] | Exhibit halls can improve client engagement if an indoor area engineering is designed and tried. | BTLE | | Indoor | | X | X |
| [53] | The proposed design involves storing clients' interactive media content in the Cloud and distributing it through the clients' informal organization as events. | PaaS & IaaS | X | Indoor/Outdoor | | | X |
| [54] | The Smart Building plan was discussed as a creative principal motor to address its issues. | ZigBee | X | Indoor/Outdoor | X | | |
| [55] | A science fiction prototyping (SFP) scenario recommendation for | SFP | X | Indoor | | | |

| Authors | Goal | Protocol | Efficiency | Surveillance environment | Security | Speed of transmission | Service quality |
|---|---|---|---|---|---|---|---|
| | smart structures based on morphogenetic plans | | | | | | |
| [56] | Clarifications concerning the AI-based smart building automation controller (AIBSBAC) designed and organized using AI. | AIBSBAC | | Indoor/Outdoor | | X | X |
| [57] | In the future, temperature controls in smart buildings will be based on detailed thermal models developed using the IoT | Bluetooth | | Indoor | | X | |
| [58] | Discussion of current state-of-the-art technology development, parameter management, and IoT infrastructure needed for smart development. | Wi-Fi | | Indoor/Outdoor | | | |
| [59] | Presented current cutting-edge exercises that play a pivotal role in shrewd structuring with innovative programming, boundary setting, and the use of IoT. | Z-WAVE | X | Indoor/Outdoor | X | | X |
| [60] | The smart city era was discussed in terms of its main results, technical challenges, and socio-economic benefits. | SFP | | Indoor | | X | |
| [61] | IoT technology may be used to regulate the performance of all physical systems in smart buildings to achieve energy efficiency. | LTE | X | Indoor/Outdoor | | | |
| [62] | New matrix referred to as the "Laplacian IoT matrix" has been developed that provides information on | Wi-Fi | | Indoor/Outdoor | | | X |

| Authors | Goal | Protocol | Efficiency | Surveillance environment | Security | Speed of transmission | Service quality |
|---|---|---|---|---|---|---|---|
| | connected devices and smart buildings. | | | | | | |
| [63] | This framework introduces key concepts of IoT as they apply to smart homes so that we may understand and use them. | Bluetooth | X | Indoor | | | |
| [64] | Building a control system based on the IoT architecture | MQTT | X | Indoor | | | |

We explore various studies exploring the integration of IoT technologies into energy management systems within smart buildings and cities in the literature review. These studies demonstrate the transformative potential of IoT-based solutions. These works reveal key themes, including the implementation of machine learning and big data analytics to optimize energy consumption, the creation of sophisticated algorithms for energy management in the home and public sector, and the development of smart city architecture. Through intelligent analysis and real-time data processing, these investigations demonstrated how IoT technologies can effectively reduce energy consumption and environmental impact. Literature reveals a trend toward IoT framework adoption that facilitates energy management and promotes the development of third-party applications, thus enhancing the adaptability and utility of smart systems. Scalable, flexible solutions that can adapt to changing energy demands and technological advancements are becoming increasingly important. The critical analysis of these studies places the research within a broader context of IoT applications. As well as the challenges and opportunities that lie ahead, it outlines innovative approaches and technological advancements. Comparison and contrast of previous research can help us better understand how IoT-based systems can address energy consumption and sustainability issues in smart environments. IoT capabilities and their impact on smart city infrastructure are examined in this study, reinforcing the significance of the study and guiding future research. In the study, we utilized an analytical approach to assess the effectiveness of IoT-based frameworks for energy management in smart cities. The analysis began with descriptive analytics to summarize baseline energy usage from IoT device data, followed by predictive analytics using machine learning algorithms like regression analysis and neural networks to forecast future energy demands. We then applied prescriptive analytics with optimization models such as linear programming to suggest actionable energy management strategies. Simulation models tested the feasibility of these strategies under various conditions, while data fusion techniques integrated diverse data sources for an energy overview. Additionally, system dynamics models assessed long-term impacts on urban infrastructure, and real-time monitoring and control ensured optimal system performance by responding immediately to any changes in energy usage. Each method played a critical role in providing a thorough understanding of the operational efficiencies and strategic decision-making necessary for sustainable urban development.

### *2.1. Components of smart cities*

The components of a smart city are shown in Figure 1. A smart city application starts with data gathering, then transmits and receives data, stores data, and analyzes the data. In several sectors, sensor development has been driven by the requirement to collect data. The second step is the transmission of data from the data collection devices to the cloud for analysis and storage. There have been many smart-city initiatives that have included citywide Wi-Fi networks, 4G and 5G technology, as well as local networks that can send and receive data locally or globally. In preparation for the fourth step, data analysis, cloud storage is the third phase, which incorporates numerous storage options. The analysis of data helps in making decisions by finding patterns and conclusions.

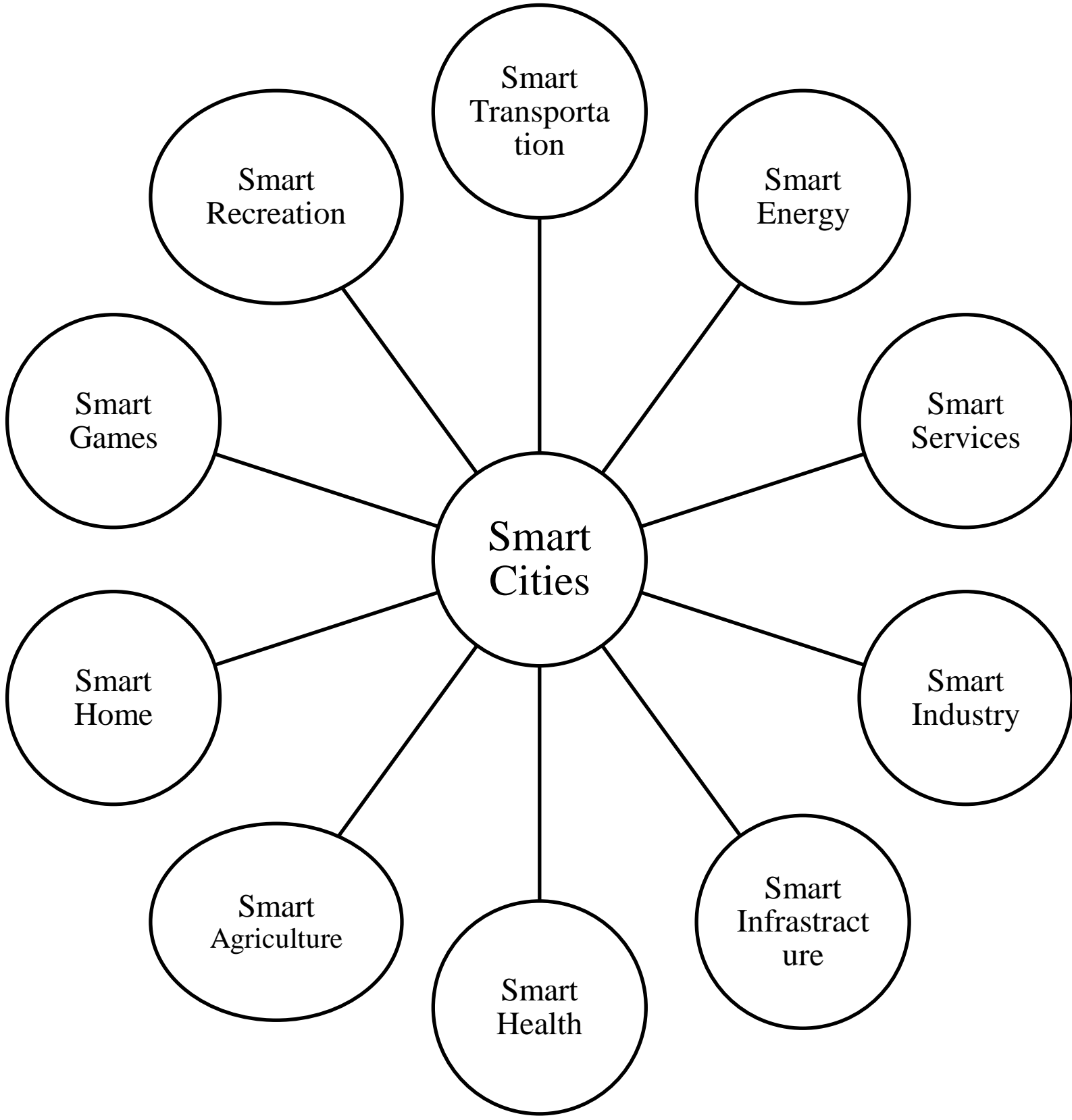

.

**Fig. 1.** Key Components of Smart Cities: This diagram illustrates the interconnected elements that constitute smart cities, including Smart Transportation, Smart Energy, Smart Services, Smart Industry, Smart Infrastructure, Smart Health, Smart Agriculture, Smart Home, Smart Games, and Smart Recreation, highlighting the holistic approach to urban development.

There may be circumstances in which simple analyses, such as decision-making and aggregation, are appropriate. Cloud computing makes it possible not only to gather, save, and process data, but also to analyze it in real-time utilizing deep learning algorithms and machine learning algorithms [6] (see Table 2).

**Table 2**: Components of smart cities

| Author | Year | Method | Purpose/Result |
|---|---|---|---|
| Chen et al. [65] | 2023 | smart city quality of life (SCQOL) | Using the five SCQOL domains, their impact on citizens' support for smart city development (SCD), and the three identified citizen segments, the necessary actions can be implemented to improve SCD for the target groups. |
| Weber- | 2023 | digital twins and AI model | By employing digital twins and artificial intelligence, |

| Lewerenz [66] | | | it is possible to create advanced fire prevention systems that utilize specialized automated building solutions. These systems can be customized, tested, planned, and modified according to specific needs, and can effectively communicate important information through visual representations. It can be said that a single image generated by these systems can convey a vast amount of information, equivalent to a thousand words. |
|---|---|---|---|
| Chen et al. [67] | 2023 | sustainable development goals (SDGs) | The availability of green spaces, recycling initiatives, and measures to mitigate air pollution are connected to happiness. When it comes to enhancing people's wellbeing, urban infrastructure plays a more crucial role than technological gadgets. |
| Savastano et al. [68] | 2023 | information and communications technology (ICT) | According to the results, institutions and providers of smart mobility solutions must have a clear comprehension and effective communication of how digital services can be utilized at different user touchpoints and communication channels to enhance the value provided to both residents and visitors. This is essential to improve the overall experience of the users. |
| Buhalis et al. [69] | 2023 | A total of 145 peer-reviewed publications on smart hospitality were collected from Web of Science. Additionally, eight recent reviews on smart tourism and hospitality were examined to provide a analysis of the topic. | Drawing from prior systematic reviews of smart hospitality and literature evaluations, this study aims to investigate the latest trends, issues, and themes in this field. It consolidates existing knowledge, extrapolates insights, and contributes to the progress of smart hospitality by serving as a valuable resource for stimulating academic-business discussions and inspiring further research. |
| Kuo et al. [70] | 2023 | Internet-of-Things (IoT) | This study will address the challenges encountered by smart public transportation systems in areas such as network design, operational planning, scheduling, and management. |
| Zeng et al. [71] | 2023 | the q-rung orthopair fuzzy set (q-ROFS), TOPSIS model | The aim of this paper is to present a novel method for assessing q-ROF (Quality of Resilient Operation Framework) scenarios, which involves an advanced version of the TOPSIS (Technique for Order Preference by Similarity to Ideal Solution) model along with specific operators. The proposed technique is used to evaluate a smart city, and the results demonstrate the feasibility and effectiveness of the system. |

| Verhulsdonck et al. [72] | 2023 | A Systematic Review | This study aims to assist smart city planners in developing and evaluating cybersecurity measures that incorporate both personal privacy and engaging features. To achieve this objective, the research proposes using a cybersecurity lens, such as the McCumber cube model, which can provide a framework for analyzing various aspects of cybersecurity. |
|---|---|---|---|
| Alshamaila et al. [73] | 2023 | A systematic review | This study provides an in-depth examination of each issue and offers recommendations for further research. |
| Twist et al. [74] | 2023 | A systematic review | This study's systematic review focuses on citizen dissatisfaction with smart cities, which can be viewed in two ways: active and passive. Additionally, the research examines government efforts to address and promote critical citizen engagement. |

### *2.2. IoT definition in smart devices*

Intelligent devices will combine sensor data from several sources, determine the best course of action based on local and distributed information, and then change or manage the environment. By utilizing computing resources, storage, and interlayer communications, such as cloud computing, cloud storage, and other current Internet technologies, objects can operate in the physical realm and provide enhanced robotic services. Cloud data transmission speeds provide robot mobility and functionality, allowing jobs to be transferred without severe restrictions. Robotics is undergoing rapid change with the emergence of communication-oriented robots that utilize wireless connections and links to sensors and other network resources. Robots connected to a network are robots that interact with the Internet or a LAN through a communication network. There are multiple ways to create networks, including wireless, wired, or using protocols such as TCP, UDP, 802.11, and more. Numerous new possibilities are being explored with these robots, including automation and exploration. The Association of Automation and Robotics in Network Robots defines two sub-categories of network robots (network-connected) :Tele-operated robots are robots to which the human user sends commands and receives feedback through the network. Such systems are used in research, education, and public awareness using valuable resources available to a wide audience.

With autonomous robots, robots and sensors exchange information over a network with minimal human intervention. The sensor network expands the effective measurement range of robots and allows them to communicate with one another over long distances to coordinate their activities. Likewise, the robot can assist in the deployment, maintenance, and maintenance of the sensor network to extend its life and usefulness In both groups of network robots, a common challenge is developing a scientific base that can be used to link control and create new capabilities. The typical robot is a high-capacity closed system in which upgrades or changes in performance and operation (remote or local) require extensive knowledge and are usually subject to long maintenance periods, and possesses no free communication interfaces, and this is usually done to ensure security and control the effectiveness of the robot. The concept of the IoT in intelligent robots goes beyond network and mass robotics, which integrate heterogeneous intelligent robots into a distributed architecture of cloud-and-edge operating systems. "IoT in intelligent robots" offers intelligent and advanced robots with high and new capabilities by integrating IoT and intelligent robots in today's technology. The development of multi-radio access technology to connect smart devices to the edge has created heterogeneous cellular networks that require complex configuration, management, and maintenance to support intelligent robots. The use of artificial intelligence techniques enables IoT-enabled robotic systems to integrate seamlessly with IoT capabilities to create optimal solutions for specific applications. IoT technologies in robots enable the embedding of information in systems and processes, enabling companies to increase efficiency, identify new business opportunities, and as a result, IoT-based robotic systems are better prepared to meet multiple needs within more complex environments [28], as exhibited in Figure 3 and Table 4.

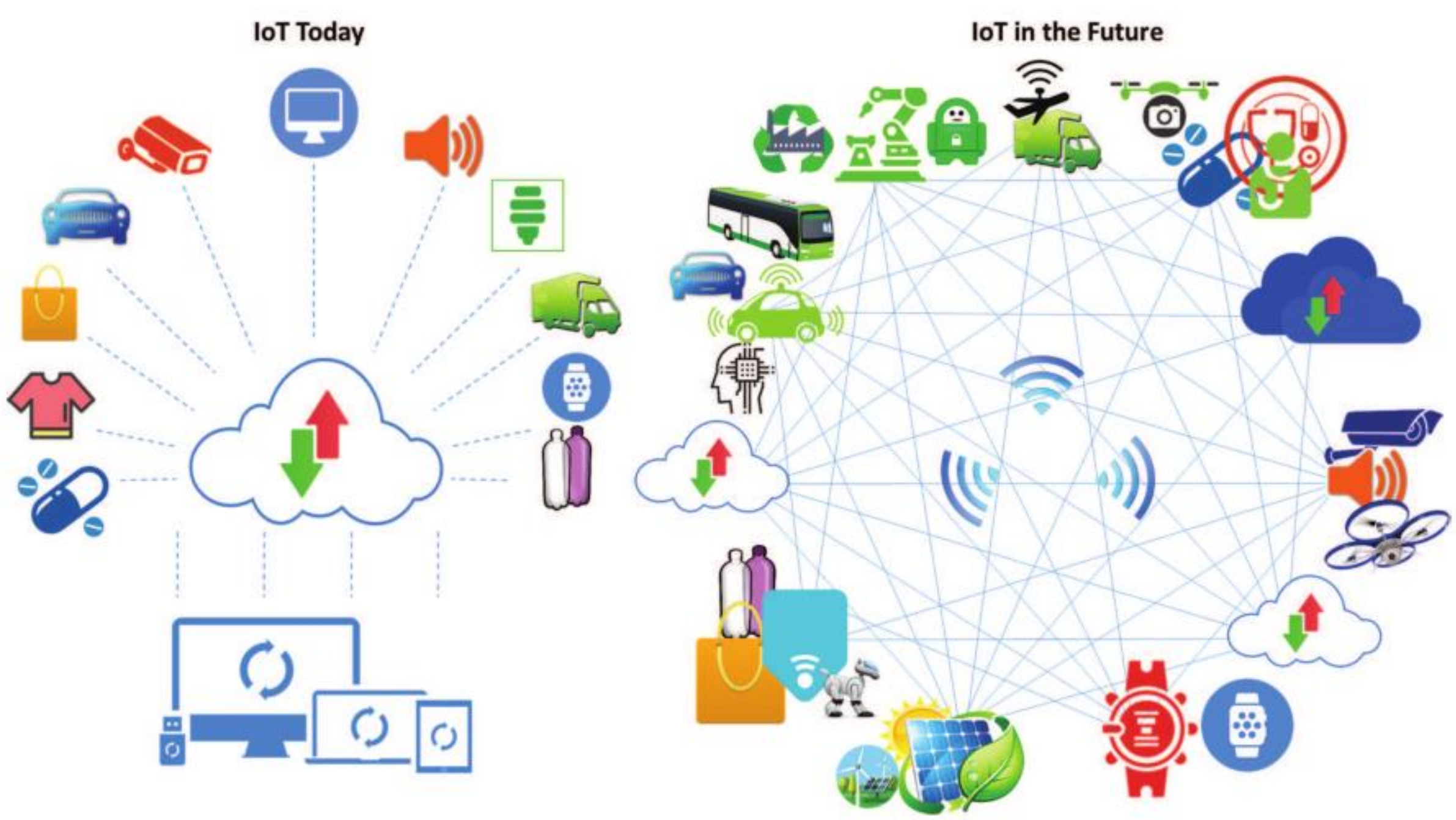

**Fig 3:** IoT now and IoT in the future through smart robots

**Table 4:** Works related to IoT definition in smart devices

| Author | Year | Method | Purpose/Result |
|---|---|---|---|
| Luperto et al. [85] | 2023 | Internet of Robotics Things (IoRT) | This research proposes a method to integrate and establish synergy between the IoT system and robots, and demonstrates how this approach can enhance long-term care for elderly individuals. The study shows that the combined use of IoT and robots can result in improved performance and effectiveness in providing care. |
| Prajapati et al. [86] | 2023 | IoT | The deployment of IoT devices on farms can enable data collection and processing at frequent intervals, allowing farmers to respond promptly to emerging problems and environmental fluctuations. This can streamline the farming process and lead to improved outcomes. |
| Pouresmaieli et al. [87] | 2023 | IoT and sustainability challenges | The study's results revealed that the implementation of IoT in mining activities can contribute to the economic index of sustainability by increasing GDP, wealth, income, and productivity while simultaneously reducing total cost, operating costs, and depreciation costs. |
| Kasturi et al. [88] | 2023 | IOT-Based Child Safety Monitoring | The proposed kid safety monitoring system in this research utilizes low-cost and easily accessible electrical components, |

| | | Robot with User-Friendly Mobile App | resulting in a cost-effective IoT-based solution for monitoring child safety. |
|---|---|---|---|
| Ryalat et al. [89] | 2023 | cyber-physical systems (CPS), IoT | This study outlines the development of a smart cyber-physical system that aligns with the innovative Industry 4.0 framework for intelligent factories. The system integrates the essential industrial, computer, information, and communication technologies of a smart factory. |
| Meddeb et al. [90] | 2023 | Raspberry-PI and IoT | This study evaluates the performance of the proposed face recognition algorithm and compares it with existing approaches. The analysis includes a set of numerical results for comparison purposes. |
| Cascalho et al. [91] | 2023 | MQTT protocol, IoT | Integrating IoT and robotics into the classroom can promote the development of students' creativity and enable them to undertake projects in key domains related to IoT and robotics knowledge. |
| Sadiku et al. [92] | 2023 | IoT | The manufacturing industry has undergone transformations and progressed in terms of automation, computerization, connectivity, and technological sophistication. |
| Yadav et al. [93] | 2023 | Raspberry-Pi, IoT | The IoT-based construction of the robot is intended to utilize a camera for environmental observation and a Raspberry Pi for monitoring its condition. |

### *2.3. Smart city IoT devices*

Smart cities are driven by the IoT; it is the technology that enables ubiquitous digitization, thus providing the basis for smart cities. Objects are being connected to the internet more and more, enabling them to transmit data and receive commands. This is known as the IoT. With the IoT, you can collect data and use data analytics to provide useful information for policy and decision-making. The Internet is expected to connect 75 billion devices by 2025 [7], allowing even more applications to be created. Using the IoT, sensors in smart cities could collect and send data about the condition of a city, which could then be analyzed and used for decision-making. In terms of big data and intelligent traffic signal systems, the classification of data positions is of critical importance. The data in the corners correspond to the districts on Google Maps [8]. The data also includes a time column that indicates congestion length. As a result, it will be possible to determine the length and location of daily congestion, as well as when and where it occurs. It is impossible to overestimate the importance of IoT big data management surveys and analytics. Figure 2 depicts an all-encompassing perspective. Among the sections of the IoT are Big Data Management, Analytical, Security, Privacy, and Energy Efficiency [10]. In addition to categorizing and storing data in a database, the big data framework must also place special emphasis on unstructured data. Scaling big data is another crucial function for reducing data storage size (see Table 3). Deep learning and analytics are two methods of examining massive amounts of IoT data. IoT creates big data that is unparalleled in its size, variety, and significance. Structured, semi-structured, and unstructured data exist.[11].

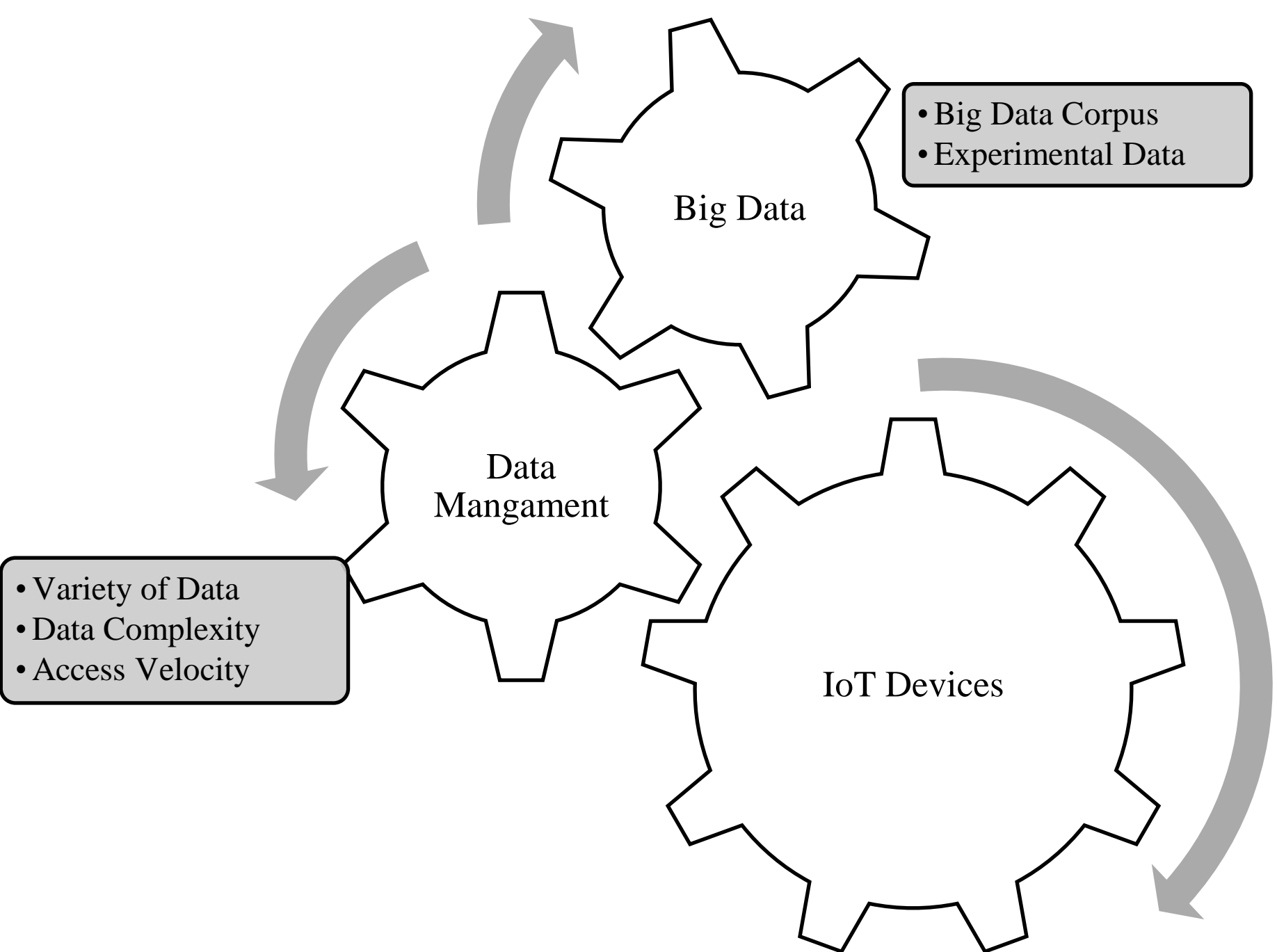

**Fig. 2.** Managing big data through IoT devices.

**Table 3:** Works related to smart city IoT devices

| Author | Year | Method | Purpose/Result |
|---|---|---|---|
| Ahmed et al. [75] | 2023 | smart Enterprise Management System (EMS), Augmented Intelligent Telemedicine (AITel) framework | The proposed architecture for eHealth services is based on a smart Enterprise Management System (EMS). The Augmented Intelligent Telemedicine (AITel) paradigm, which leverages augmented intelligence to enhance telemedicine, is suggested to achieve an accuracy rate of 94.83%. Additionally, it is envisioned to create a reliable ecosystem for resource recommendations, thereby facilitating the establishment of a resilient healthcare system for remote patients, communities, and infrastructure. |
| Lucic et al. [76] | 2023 | An Overview | The purpose of this study is to provide an overview of the practical applications that micro-unmanned aerial vehicles (UAVs) can offer to smart cities, with a focus on intelligent transportation systems (ITS). The research also identifies potential challenges associated with UAV operations and suggests several feasible approaches to managing UAV swarms while addressing the issue of their limited battery capacity. |
| Kumar et al. [77] | 2023 | IoT, V-model development lifecycle levels | By identifying the challenges encountered in different projects and categorizing them based on the V-model development lifecycle, this study provides a reference guide for future projects. Early recognition of these issues can aid ongoing and upcoming IoT research initiatives in smart cities to accelerate testbed delivery and reduce implementation time. |
| Majid et al. [78] | 2023 | IoT | This journal article investigates the essential features of smart cities and the security prerequisites for IoT systems that facilitate them. It also delves into the major privacy and security issues associated with the application architecture of smart cities. |
| Patil et al. [79] | 2023 | IoT | The objective of this article is to provide a review of the latest research on IoT in smart cities, while also identifying potential areas for further investigation. |
| Ali et al. [80] | 2023 | Generic Middleware for Smart City Applications (GMSCA), IoT | The performance and load balancing tests are employed to assess the effectiveness of the GMSCA (Global Mobile Satellite Communications Antenna) system. The results indicate that the GMSCA operates efficiently and is highly functional. |
| Chithaluru et al. [81] | 2023 | Low-energy Adaptive Clustering Hierarchy (LEACH) and Low-energy Adaptive Clustering | After analyzing the results, it is observed that the proposed protocol outperforms LEACH and LEACH-C, demonstrating an average improvement of 35% in key performance metrics such as First Node Dies (FND), Last Node Dies (LND), the |

| | | Hierarchy-Centralized (LEACH-C), IoT | number of packets sent to CH & BS, network convergence time, network overhead, and average packet delay. |
|---|---|---|---|
| Walczak et al. [82] | 2023 | IoT, structural equation modeling | The unique contribution of this study is its proposal that the positive relationship between the adoption of IoT devices and their perceived usefulness for individuals with disabilities is enhanced and moderated by the level of empathy towards people with disabilities. |
| Wen et al. [83] | 2023 | IoT, Edge computing, Distributed systems, Big data, Self-organized Maps | This study has established a framework for processing vast amounts of data generated by smart city health monitoring and diagnosis applications across geographically distributed edge clusters. The proposed architecture is founded on the Map-Reduce methodology for distributed processing of large datasets and utilizes edge clusters located throughout the smart city. |
| Zeb et al. [84] | 2023 | radio frequency (RF)-based Energy harvesting, IoT devices | The researchers utilized various techniques to ensure a continuous energy supply for the gadget, preventing it from remaining idle. These strategies comprise of probabilistic charger placement, fixed charger placement (single/multiple), mobile charger placement (single/multiple), and set covers. However, some limitations of these approaches include casting and charging times. |

### *2-4-IoT application in smart city infrastructure*

The IoT can be used in urban infrastructure such as bridges, subway lines, railways, streets, etc. The IoT can also be used in the service areas of cities and by coordinating different urban systems to do this more effectively and efficiently for people.

## 3- Communication technologies

The IoT in intelligent robots requires new communication architectures and equipment to operate effectively in the real environment, which can perform complex computing and information exchange in both the internal communication and edge computing in the real environment and can virtualize Give the best way to do it. In this regard, communication technology plays an important role in the IoT in robots.

**Table 5:** IoT layers in intelligent robots

| Layer | Protocol |
|---|---|
| Physical layer | Devices/tools / objects |
| Link layer | Bluetooth, Wi-Fi, WiMAX, ISA 100.11n, LTE-MTC, GSM, NB-IoT, Eddystone |
| Internet layer | IPv6, uIP, NanoIP, 6LoWPAN |
| Transfer layer | CoAP, MQTT, XMPP, AMPP, AMQP, LLAP, DDS, SOAP, UDP, TCP, DTLS |
| Application layer | REST API, JSON-IPSO Objects, Binary Objects |

Various IoT communication protocols are used in intelligent robots to facilitate the layer-by-layer exchange of information. Table 5 shows the different types of communication technology in "IoT in intelligent robots".

### *3-1-Advanced IoT solutions for smart grids and smart city*

Smart grids that incorporate IoT provide a fresh perspective on power management, which is advantageous to all stakeholders. Smart grids can be classified in various ways based on various components (aspects). Using the smart grid's first three elements (production, transmission, and distribution) has presented several challenges to pioneering research in this area, especially regarding production, transmission, and distribution. The harsh environment in which the sensors are located is largely to blame for the difficulties. Experimental results obtained from the use of standard sensor networks compatible with IEEE 802.15.4 show that wireless communication (includes both LOS and NLOS scenarios) in intelligent networks due to electromagnetic interference, equipment noise, Blockage, etc., closed error rate, and variable link capacity is high. Additional limitations are imposed by wireless nodes, such as memory limitations, processing limitations, and inadequate power sources (see Table 6). With the IoT, people can design intelligent solutions and bridge this chasm. The IoT is aimed at creating a better and safer society where "everything is a service" (public safety, health care, manufacturing, etc.). This section reports on relevant attempts in the scientific literature. They are mainly devoted to wireless sensor networks, home automation, and smart grids; they come under the topic of IoT solutions for the smart city.

**Table 6**: Smart grid IoT applications

| Consumers | Distribution | Transfer | Energy Providers and Production |
|---|---|---|---|
| Automatic and wireless size reading (smart measurement) | Monitoring of underground cabling system | Transmission line control | Immediate monitoring of production |
| Home (residential) supernatural energy | Control of transfer stations | Power monitoring | Control of power plants |
| Managing solar panels | - | - | Control of alternative energy sources |
| Predicting future solar panels and generating wind turbines (using sensor data such as temperature or humidity) | - | - | Monitoring the creation of a residence (distributed) |

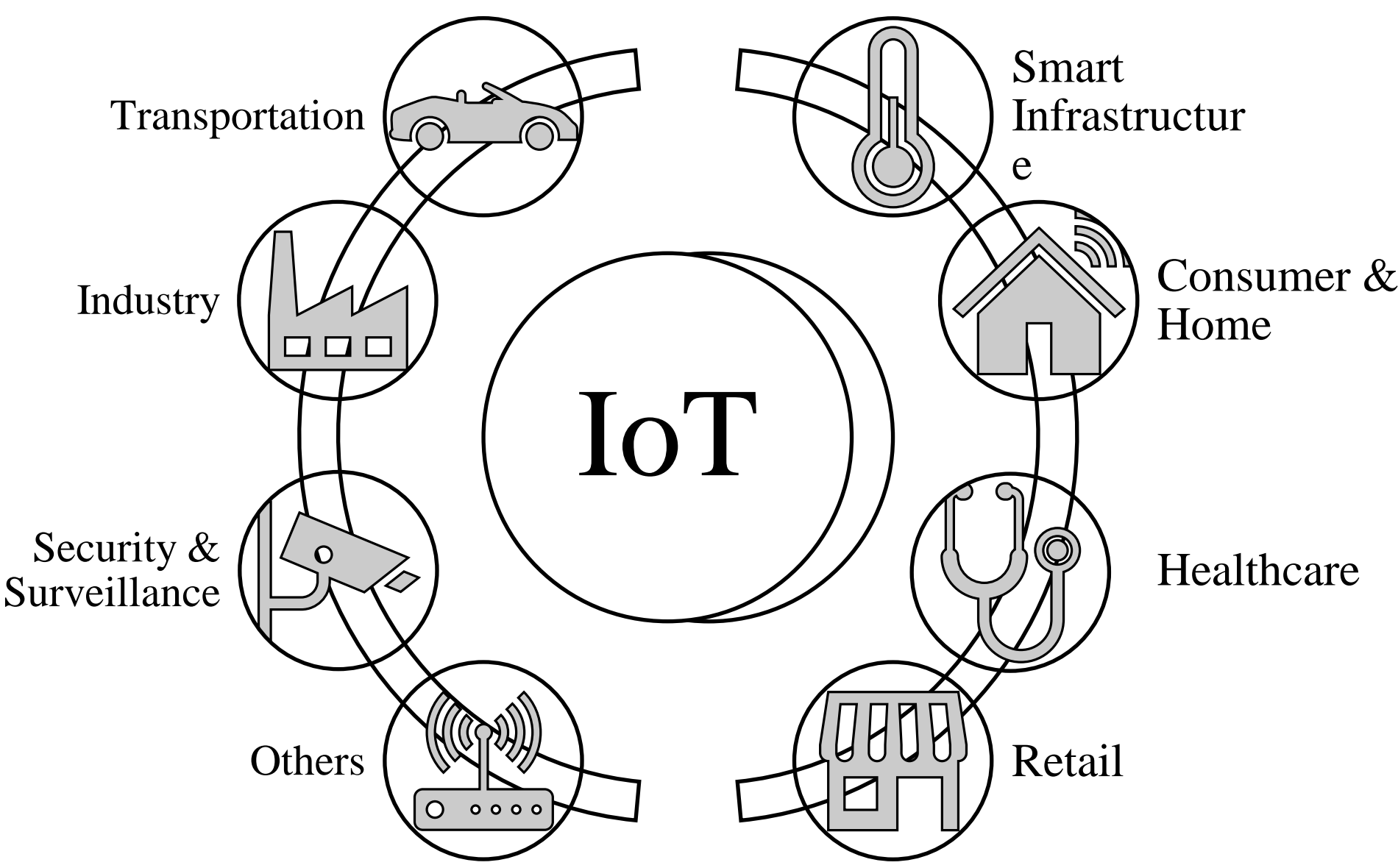

**Fig. 4**: IoT application high-level view

As a result of sophisticated software technologies, new designs have been presented that emphasize the indoor environment and habitat monitoring (See Figure 4).

## *3-2-Smart City*

An intelligent summer home has been considered for the ECOGRID EU project. The authors implemented a modular, scalable wireless sensor network in their VILLASMART experiment. The energy consumption of the building was modeled. By incorporating wireless sensor network interior and outside studies (air and water temperatures, solar radiation sensors, weather conditions, and energy consumption data), these thermal models are enhanced, thus providing more accurate projections of inside temperatures. Typical capacitance-resistance models forecast an inaccuracy of 1.790 ° C. A 2.4 GHz frequency is used for IEEE 802.15.4 internal communications. Gray estimation is used to determine the model's parameters. An energy management system connects smart devices and software that calculates energy usage. As a result, welfare services and consumers stand to gain from this approach. The use of demand management systems is another prevalent statistic. In future energy systems, these technologies will balance electricity consumption/production at the customer level. In IoT solutions designed for smart homes, the old DSM approach is giving way to cloud-based strategies. It is predicted that the cloud-based strategy will perform better than a conventional technique in energy management because it offers centralized optimization. Centralized optimization considers a set of factors, so it will consider a set of criteria. This figure demonstrates the smart urban management approach used to implement a complete framework.

### *3-3- Smart city challenges*

To digitize a city, it needs to have a proliferation of sensors in every aspect of its operation. Smart cities present enormous challenges in building and implementing IoT systems because of such a vast application field. This section focuses on the challenges that IoT system designers face while deploying smart city applications. Academics have been interested in IoT adoption in smart cities because of the technical challenges involved. Figure 5 highlights the several obstacles that Smart City IoT system implementations must overcome, such as Security and Privacy, Smart Sensors, Networking, and Big Data Analytics [30] (see Table 7).

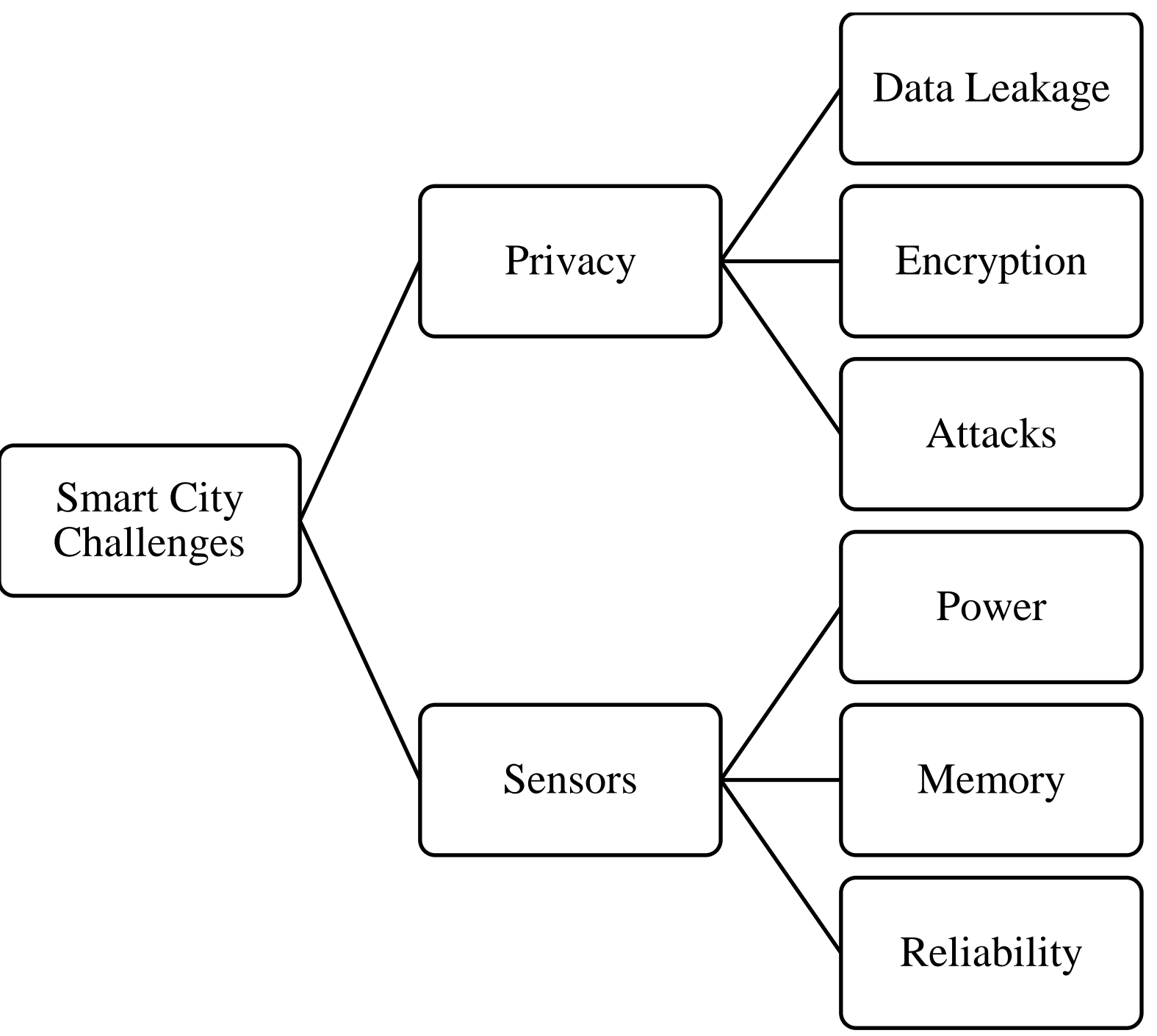

**Fig. 5.** Management of big data by IoT devices

### *3-4- About smart city privacy*

Smart cities are the most secure and private when it comes to security and privacy. It would be inconvenient and hazardous for its residents if the city's services were disrupted. Smart cities will only work if important municipal facilities are online, so security will not be an issue. Smart cities are more vulnerable to cybercrime and cyberwarfare in an era when they are a strategy in global politics. This calls for the encryption of network data transfers. Citizens' confidence and participation are essential to the success of smart city programs. With sensors continuously collecting information about people's activities, the expansion of sensors in smart cities may expose residents' daily activities to undesirable parties. Additionally, organizations and companies using the IoT network may collect and exploit citizen data without their awareness for eavesdropping or targeted advertising purposes. To allow for proper decision-making, solutions will need techniques that anonymize data collection while keeping the context of the measured job.

**Table 7:** Works related to smart city privacy

| Author | Year | Method | Purpose/Result |
|---|---|---|---|
| Rawat et al. [94] | 2023 | AI, IoT | The purpose of this study is to emphasize, particularly in the context of security, the usage and significance of drones for intelligent city management. |
| Chentouf et al. [95] | 2023 | Blockchain, Cybersecurity | The article analyzed the potential contribution of blockchain features like transparency, democratization, decentralization, and security towards improving smart city services. It also reviewed a number of blockchain applications in smart cities. The study aims to showcase how blockchain technology can enhance security in smart cities by implementing an electronic voting system through a smart contract developed on the Ethereum blockchain. |
| Siddiqui et al. [96] | 2023 | Software Defined Networking (SDN), IoT | To evaluate the practicality of the proposed security framework for services, this research created a use case that demonstrates the effectiveness of collaborative services in an IoT architecture enabled by Software-Defined Networking (SDN). |
| Ro et al. [97] | 2023 | AHP (analytic hierarchy process), IoT | The study found that effectiveness, efficiency, and sustainability are important considerations, while the security and privacy of technology (Tech 4) were the primary concerns for smart city collaboration. |
| Khanpara et al. [98] | 2023 | IoT | To identify and stop such risks, this study investigates several security dangers in a smart home setting and suggests a context-aware security-based system. |
| Shalender et al. [99] | 2023 | IoT | The framework highlights the specific areas that require specialized focus from the industry and proposes solutions to effectively address the security and privacy issues related to data. The research also offers crucial recommendations for businesses, consumers, and policymakers and has implications for both researchers and practitioners. |
| Saini et al. [100] | 2023 | IoT | The primary focus of this research is on smart cities, exploring what makes them intelligent and how they function, the challenges involved in making cities smart, the various types of threats and risks they encounter, the ongoing initiatives to enhance their safety and security, and the potential areas for further advancement. |

### *3-5- About smart sensors*

For smart cities to conclude, sensor equipment must communicate data, schedule activities, and aggregate data. By developing and accepting open protocols, manufacturers can create interoperable equipment, which will accelerate the deployment of IoT systems. It is also possible to create 'standard' access point nodes for IoT systems that can communicate with devices utilizing a wide range of communication protocols and analyze incoming data. As indicated in [31], several equipment manufacturers have enabled compatibility with alternative protocols. Reliability and durability are also concerns with intelligent sensors. The IoT system's reliability and robustness are determined by its dependability and precision. The IoT is at the heart of smart cities, and since it's so crucial, the system must provide a seamless experience for its citizens. Users' requests for service must be handled promptly and accurately. Every citizen should have access to high-quality services in the smart city (see Table 8) .

**Table 8:** works related to smart sensors

| Author | Year | Method | Purpose/Result |
|---|---|---|---|
| Chi et al. [101] | 2014 | wireless sensor networks (WSN), complex programmable logic device (CPLD) | The efficiency of the proposed system has been validated, and the practical implementation of the IoT in monitoring aquatic environments has yielded favorable outcomes. |
| Sehrawat et al. [102] | 2019 | IoT | This article describes numerous sensor-based IoT applications as well as a number of IoT sensors. This article also clarifies which sort of sensor is needed for which IoT application by studying various sensor applications. |
| Ullo et al. [103] | 2021 | AI, IoT | The article provides a detailed analysis, evaluation, and comparison of specific types of sensors and their technologies. They also suggest IoT advancements that could assist researchers, farmers, and policymakers in their remote sensing and agricultural research and implementations. |
| Martins et al. [104] | 2023 | IoT sensors | The "online lab" is an intriguing and useful resource, especially in the current pandemic situation. Additionally, it offers students the opportunity to work with tangible hardware, which can aid in enhancing their skills and understanding of IoT sensor devices. |
| Basith et al. [105] | 2023 | IoT-cloud supported | This research makes a valuable contribution to crisis management efforts by repurposing pandemic waste for energy harvesting and sensing applications. It also serves to mitigate the problem of microplastic pollution in the environment, as well as control the spread of the coronavirus through promoting proper hand washing practices. |
| Khan et al. [106] | 2023 | The Analytical Hierarchy Process (AHP) and the Multi-Objective Optimization on the basis of Ratio Analysis (MOORA) technique. | Smart cars equipped with sensors can detect emergencies and help prevent accidents that may result in injuries or fatalities. These sensors can also be used to monitor and manage activities for enhanced efficiency. |
| Mohammed et al. [107] | 2023 | GUI cross platform mobile application, IoT | The proposed system leverages the latest IoT microcontroller and hardware, which enhances the |

| | | | precision and speed of the entire system. |
|---|---|---|---|
| Chen et al. [108] | 2023 | DNN splitting framework called NNFacet, IoT | Extensive research indicates that NNFacet outperforms four baseline techniques in terms of system longevity, latency, and classification accuracy. |
| Uppal et al. [109] | 2023 | ML techniques | The results indicate that machine learning techniques applied to sensors can accurately predict faults in smart offices, with Random Forest being the most effective method, achieving a maximum accuracy of 94.27%. Deep learning has the potential to produce even more precise results by utilizing larger datasets in the future. |

## 4- A smart compact energy meter (SCEM) based on IoT

SCEMs based on IoT is designed to increase the performance, reliability, and features of existing systems by implementing an active customer strategy. The best SCEMs collect all data related to energy. Power quality is addressed, as well as the possibilities for improving DSM Remote control, easy updates, compact size, and low-cost devices are disadvantages of the conventional system. As shown in Table 1, the IoT system described in the paper compares with the hardware setup discussed in the literature.

### *4-1- Objective*

An innovative method categorizes the available load in the locality as primary or secondary to view the power consumption pattern in the area and automate the control of the load. It is considered a major load to have lights, a fan, and a USB charger, whereas a secondary load is to have an air conditioner. A Blynk application signal is sent to the controller to control the main and secondary loads.

### *4-2- Architecture of commercial building energy management system (CBEMS)*

The CBEMS includes a PZEM-004T sensor and an ESP8266 wireless module for providing real-time information regarding power consumption, reactive power consumption, power factor, voltage, current, and RMS values. In the load and wireless modules are ESP32 Microcontroller Units and ESP8266. The ESP32 microcontroller family features built-in Wi-Fi and dual-mode Bluetooth and is low-cost and low-power. This antenna switch module includes an antenna switch, an RF balun, a low-noise receive amplifier, filters, and a power-management module. Microchip with a TCP/IP stack and microcontroller, ESP8266 is a low-cost Wi-Fi microchip. This device allows microcontrollers to connect to wireless networks and establish TCP/IP communications without requiring additional hardware. Data is transmitted via Wi-Fi from the Wi-Fi modules to a Raspberry Pi 4B+ connected with a hotspot, which then forwards it via cloud Message Queuing Telemetry Transport (MQTT) to the Blynk application of an authorized customer. Before sending a report to Blynk, the Raspberry Pi Controller calculates energy usage information and tariffs. Figure 6 illustrates the CBEMS architecture.

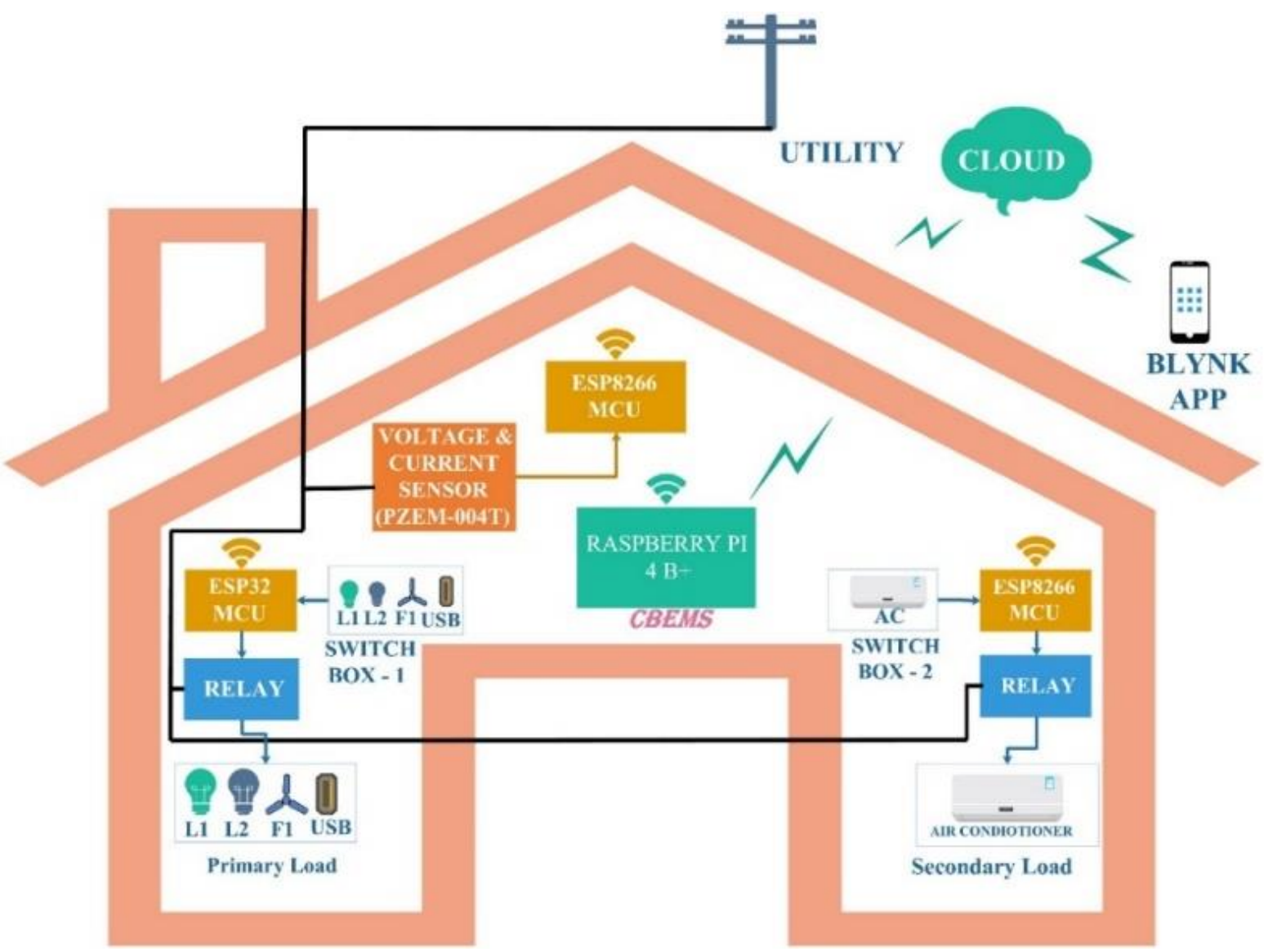

**Fig. 6:** A description of the overall architecture of an energy management system for commercial buildings.

Blynk resends information from customers to microchip modules through Blynk. When the virtual button on the User Interface of the Blynk program is set to 1/0, the relay turns the load on/off. The controller will turn off the secondary load even if the client switches it on to keep the Time of Use tariff within the nominal range. Figure 7 shows the SCEM graphically.

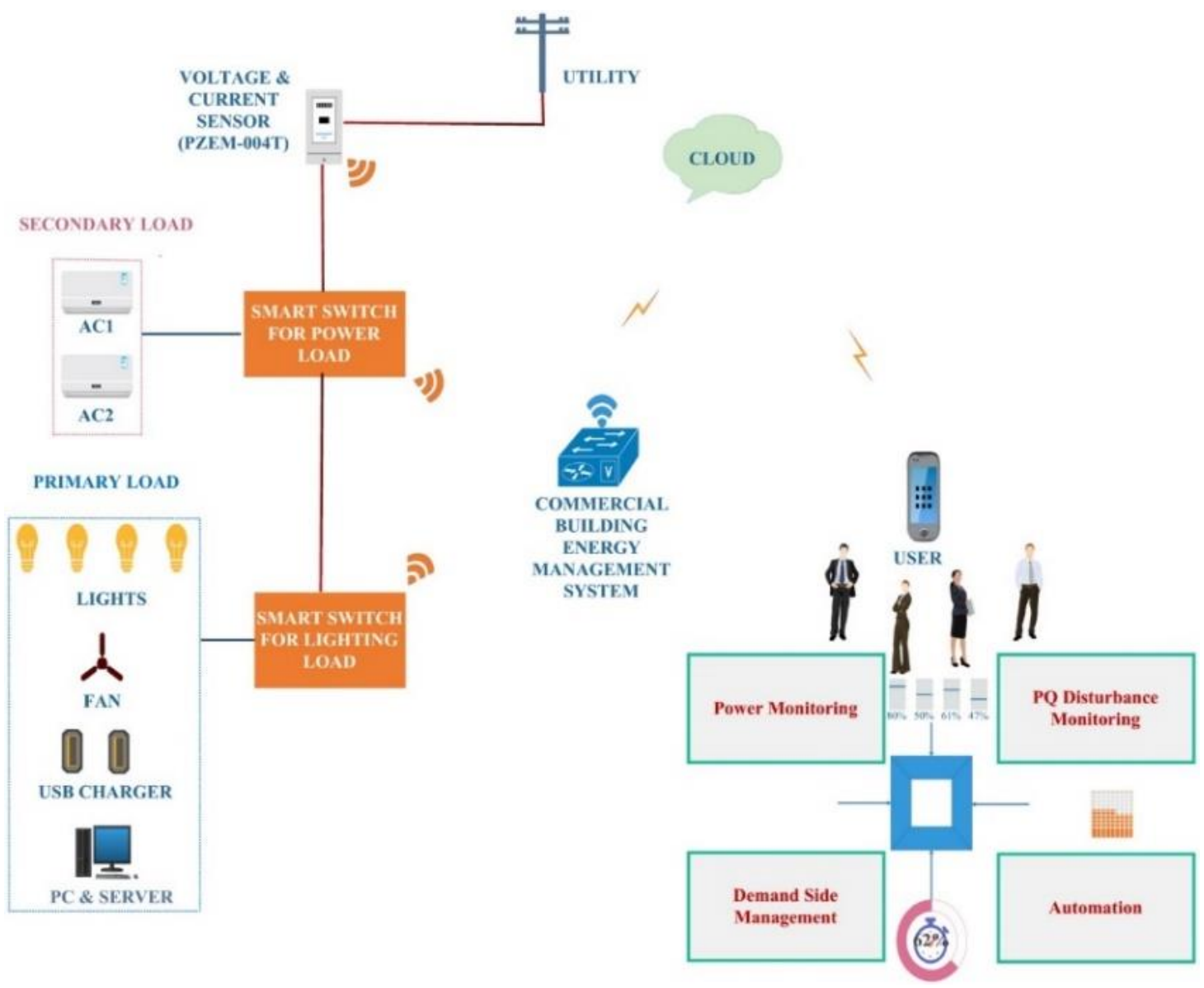

**Fig. 7.** A graphical view of the Smart Compact Energy Meter

### *4-4- Smart devices / smart tools*

Connected to production and transmission lines, intelligent objects may include appliances, lights, or sensors. These objects can sense data, activate data, process data, and communicate between data. Objects that sense and activate data must perform A/D and D/A conversions. These devices perform measurements periodically and send sensed data (either wirelessly or wirelessly) to the hub. In addition, if protocols allow, sensed data can be sent directly to the cloud. Smart devices should send the sensed data after processing the primary data. Remote data activation is also possible. According to DSM, household appliances can be divided into three categories: fixed appliances, flexible appliances, and tools that are naturally dual-use. It's not possible to manage rigid devices, such as lights, televisions, computers, or hair dryers, which relate to initial loads or other non-preventative activities. A flexible appliance is designed to handle regular loads or preventive maintenance (such as heating or cooling) and may be operated automatically. Dual-purpose instruments are certainly versatile (such as a washing machine, dishwasher, or dryer), but they can also be flexible at times. Clients may not be concerned with the exact time the washing machine operates unless the time is set within a certain time limit. These devices usually provide loads. The smart power grid allows smart devices (both flexible and dual-function) to measure and control their energy consumption in real-time(see Table 9).

**Table 9**: Works related to Smart devices

| Author | Year | Method | Purpose/Result |
|---|---|---|---|
| Harwood et al. [110] | 2014 | - | Smart-device participation, but not use, was found to be a predictor of sadness and stress, indicating that the kind of use, as opposed to the amount of use, is what mattered. |
| Lazar et al. [111] | 2015 | - | The recommendations comprise strategies for eliminating barriers, promoting adoption, and advocating for the consideration of using these devices for short-term treatments, in addition to their typical long-term use. |
| Silverio-Fernández et al. [112] | 2018 | IoT | The objective of this research is to introduce a precise and flexible concept of a smart device, which can serve as a foundation for future studies in this field. |
| Massoomi et al. [113] | 2019 | An review | "In this review, Theyconsider commonly utilized devices, evaluate the precision and dependability of the data generated by them, and explore any potential clinical applications for the gathered information." |
| Alter [114] | 2020 | - | "The objective of this article is to study the concept of smartness and to illustrate how to employ this understanding of smartness when describing, analyzing, and designing devices and systems." |
| Fazio et al. [115] | 2023 | bio-vital markers, | "The main aim of this review is to provide a survey of wearable technologies and sensor systems that are capable of monitoring patients' physiological parameters during post-operative rehabilitation and athletes' training. Additionally, They aim to provide evidence that these technologies are effective for use in healthcare settings." |
| Suh et al. [116] | 2023 | visual analogue scale (VAS), | Individuals suffering from tinnitus may benefit from tinnitus retraining therapy (TRT) using smart devices as an effective alternative. Since conventional one-on-one counseling can be time-consuming and expensive, smart-TRT may offer a more cost-effective solution with comparable treatment outcomes. |

## 5- Architectural design of the system

An intelligent building provides its users with individualized services based on the information it contains about its contents, whether it is an office, a residence, an industrial facility, or a recreation area. Buildings not only need to be efficient but also habitable and productive since they affect people's quality of life and work. As a result of the installation of building sensors and actuators, the associated cost must be balanced by the economic advantage of energy savings. It is neither possible nor practical to control the entire building. The final energy management framework should include actual sensor data on such inputs, in addition to patterns derived from data monitoring. It can therefore adapt to changing architectural situations as well as to new situations that weren't considered in the original model. This platform is comprised of three layers, which makes it suitable for a variety of smart environments, including those related to smart buildings. An IoT-based smart building is depicted in Figure 8 as a three-layer framework.

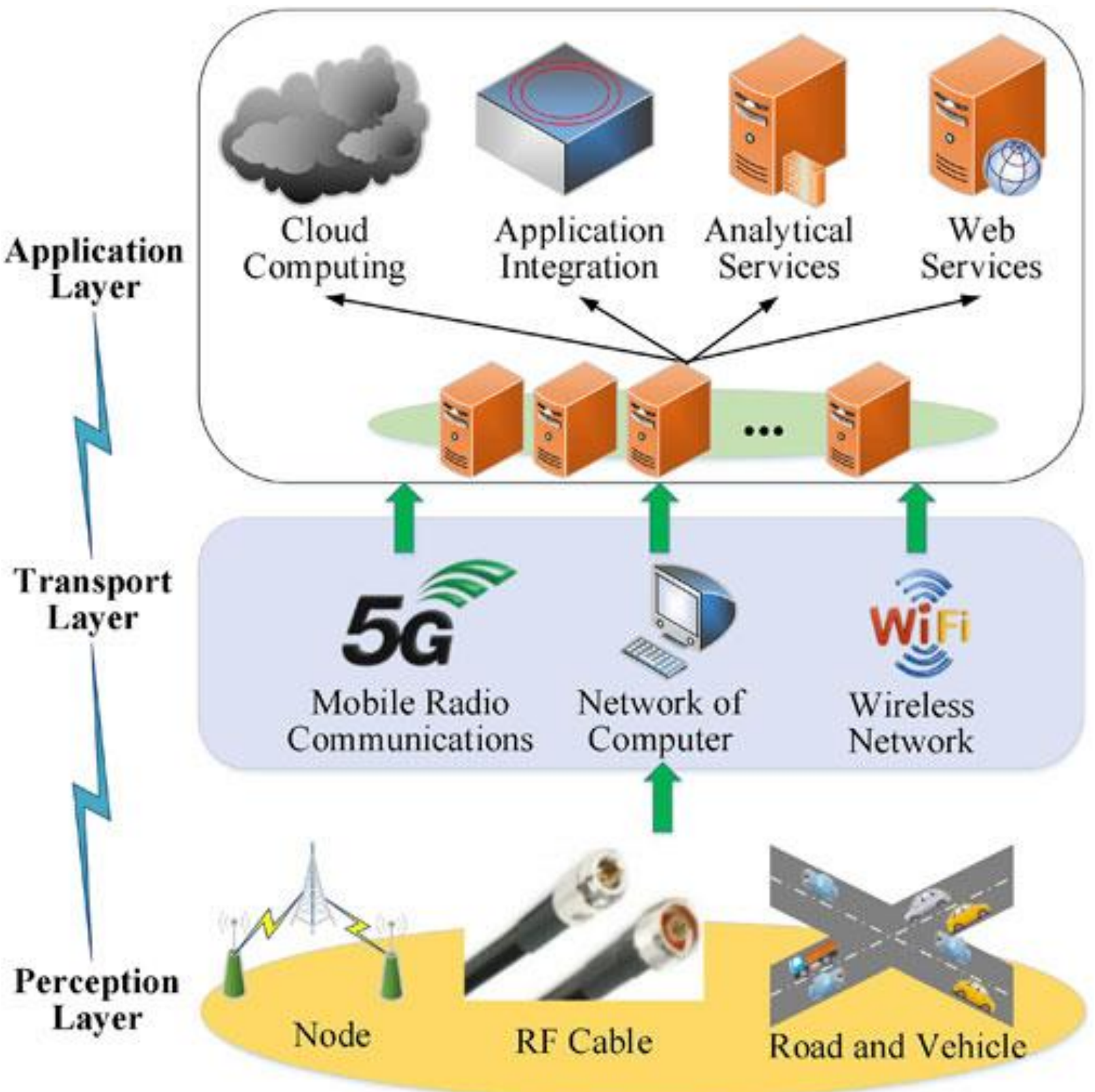

**Fig. 8.** Architecture of the system [119].

**Data Sensing or Perception Layer -** During the first layer, data is collected from sensors. Users can request information about appliances operating states, such as temperature and humidity, through this data source. A network gateway saves the information in a BD cloud.

**Data Processing or Network Layer –** The data is organized and processed in this layer. For comfort-related concerns, such as HVAC, lighting, and temperature controls, individual user data is necessary.

**Data Reproduction or Application Layer** - The third layer replicates data as a record of individual interactions between occupants and equipment. Finally, the collected data is used to improve the efficiency and functionality of the device. This leads to improved services for citizens.

In the last decade, Cloud Computing, Information Systems, and Emerging Technologies have generated amounts of data, and the miniaturization and output of IoT devices have increased. This information, however, is useless without the ability to analyze it. To extract information and make decisions, "Big Data Analysis" requires multiple levels of concentration. Humans have been able to collect relevant data using several analytical methods that integrate BD and IoT. In some ways, BD could be seen as the older sibling of IoT. As a result of all this smart technology, millions upon millions of data points are generated (IoT sensors), ushering in the age of intelligent buildings. Building upkeep and efficiency can be improved by using sensors that monitor temperature, motion, light, and humidity. Using the analyzed data from the cloud service, the motion sensor can also determine whether a resident was offering a "safe" meaning. A voltage stabilizer will also allow the mounted cloud server to operate automatically. The building's Wi-Fi connection allows users to connect via their phones to the network. By installing cameras to monitor the environment remotely from a phone or other device, intelligent technologies can be used to construct a safer building. Furthermore, the security cameras are linked to other intelligent equipment for efficient monitoring of the building based on occupancy. Through their smart devices, residents can interact with the security system. It may be connected to the lighting system, buzzers, alarms, police station, etc. A security system responds when an intruder or unusual movement is detected. There is no doubt that this intelligent defense is much more dependable and effective than an emergency siren. Additionally, homeowners will receive a maintenance plan to ensure that all equipment is functioning and operating efficiently (see Figure 8).

To monitor the well-being of elderly people living alone at home, intelligent building systems have been developed. In an intelligent building, an occupant's general physical activity, physiological processes, and environmental elements can all be monitored simultaneously. Inconspicuous, inconspicuous, and non-invasive, this device is strategically positioned throughout the structure. Providing continuous monitoring in smart buildings may be possible with just one local gateway server. Windows is used to run the software for analysis and decision-making algorithms. Access to wellness-related information is possible thanks to the Internet.

## 6- Discussion

The findings of this study illustrate the potential of IoT-based frameworks to transform energy management in smart cities and buildings. Through an analysis of existing literature, it is evident that IoT technologies can improve energy efficiency, reduce environmental impacts, and optimize operational performance. Smart cities, equipped with IoT, ML, and AI, are capable of dynamically monitoring, predicting, and managing energy consumption patterns. This capability allows cities to meet increasing energy demands while minimizing waste and environmental harm. IoT-based energy management systems offer key advantages, such as the ability to process real-time data from sensors, smart meters, and appliances autonomously. With the integration of ML algorithms, these systems can predict energy requirements, optimize resource distribution, and adapt to changing consumption patterns without human intervention. This results in enhanced energy efficiency, reduced carbon emissions, and lower operational costs. Furthermore, the inclusion of renewable energy sources, such as solar and wind, into IoT-enabled grids supports sustainable urban energy systems. Despite these advancements, the implementation of IoT frameworks still faces technical, social, and economic barriers. Technically, interoperability among devices, data security, and network reliability are persistent challenges. Socially, concerns regarding user privacy, data ownership, and equitable access to IoT systems remain unaddressed. Economically, the high cost of deploying and maintaining IoT infrastructures, particularly in resource-constrained regions, limits widespread adoption. The literature also reveals gaps in assessing the scalability and long-term impacts of IoT-based systems. Existing studies often lack standardized evaluation metrics, making it difficult to compare the performance of different frameworks. Additionally, there is a need to explore how policy interventions and public awareness can support the adoption of these technologies. Addressing these gaps is essential for advancing IoT-based energy management. Future research should focus on developing universally accepted protocols for device interoperability, robust data security frameworks, and cost-efficient solutions to make IoT technologies accessible across diverse regions. Policies to reduce the digital divide and promote inclusive adoption should also be prioritized. By tackling these challenges, IoT frameworks can fully realize their potential to create efficient, sustainable, and resilient energy systems in smart cities.

## 7- Conclusion

The Internet of Things has exhibited remarkable growth, expanding its applications beyond early focus areas such as smart homes to sectors including agriculture, industry, government, and business. Among these, smart construction emerges as a critical domain where IoT can enhance energy and operational efficiency. This study provides a review of IoT frameworks for energy management, highlighting their potential to reduce energy consumption and environmental impacts in smart buildings. IoT-equipped smart buildings demonstrate enhanced operational efficiency, safety, and the capacity to address critical needs such as comfort, accessibility, security, and energy efficiency. This paper presents an IoT-based framework for energy management in smart cities as a key contribution, that primarily focuses on a descriptive analysis wit evaluation of the framework's performance. It integrates various components of the Internet of Things architecture in order to support intelligent energy management applications, emphasizing the collection, storage, and intelligent analysis of data. Moreover, it facilitates the development of third-party applications, thereby extending its utility and adaptability.

The key findings of our analysis underscore the growing significance of IoT-based frameworks in addressing contemporary energy management challenges. While intuitive, these findings highlight critical aspects for the effective integration of IoT into energy systems:

- IoT-based frameworks demonstrate potential in improving energy efficiency and reducing environmental impacts, offering intelligent solutions to optimize resource usage in smart cities and buildings.
- The escalating demand for energy underscores the need for innovative strategies that not only enhance building maintenance efficiency but also adapt to the increasing complexity of energy systems.
- There is a pressing need to delve deeper into IoT-based wireless sensing systems, particularly their application in smart buildings. Emphasizing intelligent analysis and fostering third-party application development within the IoT ecosystem can address evolving energy demands and operational challenge.

For future research, we recommend:

- A study of AI-based energy management systems in smart cities is being conducted.
- Using IoT to manage secure, blockchain-based devices in smart cities.
- IoT applications in smart grids and environments as an energy Internet.
- Analyzing the management of IoT applications in smart environments.
- Identifying factors that facilitate the development of smart city information systems through the use of IoT and cloud computing.
- Implementing IoT-based smart applications in smart cities.

It is hoped that these directions will serve as a blueprint for researchers, planners, architects, and stakeholders who wish to integrate smart sensing technologies into future building designs in order to achieve greater environmental and energy efficiency.

## Limitation

While this study presents a review of IoT-based frameworks for energy management in smart buildings and cities, several limitations should be acknowledged. One key limitation is the reliance on existing literature and secondary data sources, which may not capture the latest advancements in IoT technologies or the evolving energy management needs in real-world applications. As IoT systems and machine learning algorithms develop rapidly, the frameworks reviewed in this study might become outdated, potentially limiting the study's applicability to future developments. Another limitation arises from the diverse range of IoT architectures and energy management systems reviewed. Although this breadth provides a wide perspective, it also poses challenges in comparing different frameworks that use varying technologies, protocols, and objectives. This diversity makes it difficult to establish a unified model or set of best practices applicable across all smart cities and buildings. The variation in technical specifications, geographical locations, and energy consumption patterns further complicates a standardized approach to intelligent energy management. Additionally, the study focuses primarily on technical aspects, such as data collection, storage, and intelligent analysis, while underemphasizing the social and policy dimensions that are critical for the large-scale adoption of IoT-based energy management systems. Factors such as regulatory policies, public awareness, user acceptance, and ethical considerations related to data privacy and security are important but not thoroughly explored. Furthermore, the study does not extensively address the economic feasibility of implementing IoT systems on a large scale. The cost of deploying smart sensors, managing data networks, and maintaining infrastructure could be prohibitive for many cities, particularly those in developing regions. Future research should include cost-benefit analyses and investigate financial models that can support the widespread adoption of IoT-based solutions in energy management. Lastly, the study primarily focuses on urban areas, potentially limiting its relevance for rural or less technologically advanced regions where IoT infrastructure and energy management systems are less developed.

## Funding statement

In this paper, the authors did not receive funding from any institution or company and declared that they do not have any conflict of interest.

## Compliance with ethical standards

**Conflict of interest:** Authors have no conflict of interest.